\documentclass[sigconf,natbib=true,nonacm]{acmart}
\AtBeginDocument{%
  \providecommand\BibTeX{{%
    \normalfont B\kern-0.5em{\scshape i\kern-0.25em b}\kern-0.8em\TeX}}}

\begin{document}

\title{Technical Report: Impact of Position Bias on Language Models in Token Classification}

\author{Mehdi Ben Amor}
\orcid{1234-5678-9012}
\affiliation{%
  \institution{University of Passau}
  \streetaddress{Innstrasse, Passau, Germany}
  \city{Passau}
  \state{Bavaria}
  \country{Germany}
  \postcode{94032}
}
\email{mehdi.benamor@uni-passau.de}

\author{Michael Granitzer}
\affiliation{%
  \institution{University of Passau}
  \streetaddress{Innstrasse, Passau, Germany}
  \city{Passau}
  \state{Bavaria}
  \country{Germany}
  \postcode{94032}
}
\email{michael.granitzer@uni-passau.de}

\author{Jelena Mitrovic}
\affiliation{%
  \institution{University of Passau}
  \streetaddress{Innstrasse, Passau, Germany}
  \city{Passau}
  \state{Bavaria}
  \country{Germany}
  \postcode{94032}
}
\email{jelena.mitrovic@uni-passau.de}

\renewcommand{\shortauthors}{Ben Amor, et al.}

\begin{abstract}
  Language Models (LMs) have shown state-of-the-art performance in Natural Language Processing (NLP) tasks. Downstream tasks such as Named Entity Recognition (NER) or Part-of-Speech (POS) tagging are known to suffer from data imbalance issues, particularly regarding the ratio of positive to negative examples and class disparities. This paper investigates an often-overlooked issue of encoder models, specifically the position bias of positive examples in token classification tasks. For completeness, we also include decoders in the evaluation. We evaluate the impact of position bias using different position embedding techniques, focusing on BERT with Absolute Position Embedding (APE), Relative Position Embedding (RPE), and Rotary Position Embedding (RoPE). Therefore, we conduct an in-depth evaluation of the impact of position bias on the performance of LMs when fine-tuned on token classification benchmarks. Our study includes CoNLL03 and OntoNote5.0 for NER, English Tree Bank UD\_en, and TweeBank for POS tagging. We propose an evaluation approach to investigate position bias in transformer models. We show that LMs can suffer from this bias with an average drop ranging from 3\% to 9\% in their performance. To mitigate this effect, we propose two methods: Random Position Shifting and Context Perturbation, that we apply on batches during the training process. The results show an improvement of $\approx$ 2\% in the performance of the model on CoNLL03, UD\_en, and TweeBank.
\end{abstract}

\ccsdesc[500]{Computing methodologies~Information extraction}
\ccsdesc[500]{Computing methodologies~Neural networks}

\keywords{natural language processing, dataset-imbalance, named entity recognition, pos tagging, language modeling}



\maketitle

\section{Introduction}
Natural language processing (NLP) is an active area of research that aims to enable machines to understand and generate human language.
Transformer-based Language Models (TLMs) have been on the rise in the natural language processing field since the introduction of BERT\cite{devlin2018bert}. 
LMs have shown state-of-the-art performances across various information extraction (IE) and natural language understanding tasks (NLU) with their ability to handle long-range context dependencies in a text while not relying on recurrent mechanisms like LSTM-based models \cite{huang2015bidirectional}\cite{peters2017semi}\cite{chiu2016named}.

 \begin{figure}[t!]
    \centering
    \includegraphics[width=0.8\linewidth]{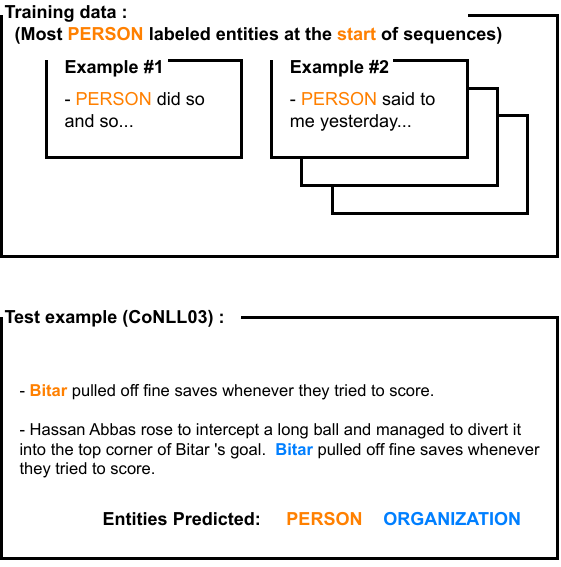}
    \caption{Example from CoNLL03. BERT model predictions on a both the original test example and an altered version. The wrong label \textbf{ORGANIZATION} was predicted for \underline{Bitar} when another sentence preceded the original example.}
    \label{fig:example}
\end{figure}

Fine-tuning pre-trained LMs for NLP downstream tasks, such as Named Entity Recognition (NER) and Part-of-Speech tagging (POS), can be challenging due to biases, class imbalances, and noisy labels that are prevalent in text datasets and benchmarks.
Named entity recognition is the task of labeling and extracting spans of named entities in text sequences.  Commonly used datasets and benchmarks in NER such as CoNLL03 \cite{sang2003introduction}, and OntoNotes5.0 \cite{weischedel2012ontonotes} suffers from data imbalances. For example, the ratio of 'None' or \textit{O} labeled tokens is 5 times as many for CoNLL03 and 8 times for Ontonotes5.0 \cite{li2019dice}. The imbalance also extends to positive class labels when dealing with more fine-grained annotations \cite{nguyen2020adaptive}. This leads to generalization performance degradation when testing on unseen data.

Another related downstream task is Part-of-Speech tagging, i.e., the process of labeling a word in a sequence as corresponding to a particular tag (called part-of-speech), based on its definition and context. The Penn Tree Bank (PTB) dataset \cite{penntreebank} is a widely used benchmark for POS tagging tasks. However, it suffers from a significant imbalance in the class distribution of the POS tags. This imbalance also extends to other benchmarks like Universal Dependencies English Web Tree (UD\_en) \cite{silveira14gold} and TweeBank \cite{liu-etal-2018-parsing} which we focus on in this study.

In addition to data imbalance issues, another challenge for LMs in NLP is position bias. So far, there have been studies \cite{ko2020look}\cite{hofstatter2021mitigating} on this bias for Question Answering (QA) tasks, where the position of the answer with the context introduces an unwanted bias when training on benchmarks like SQUAD \cite{rajpurkar2016squad}. To the best of our knowledge, the impact of position bias on token classification, like NER and POS tagging, has received less attention. 

In this paper, we define position bias in the context of token classification and demonstrate that BERT models trained on data where classes have a skewed position distribution within text sequences can be biased. It is also worth noting that an implicit correlation between context sequence length (i.e., short vs long) and position bias of tokens. Figure \ref{fig:example} shows an example of this position bias, where BERT was trained on the original CoNLL03 training set and evaluated on two concatenated test examples. The same token \textbf{Bitar, originally labeled as \textit{PER}}, was labeled as an \textit{ORG} when appearing in the middle of a longer sequence. 
To investigate the cause of this bias, we analyze the characteristics of both NER and POS datasets and study the impact of class position distribution on performance, primarily focusing on BERT models with different position embedding techniques. Position embedding (PE) gives the model information about the position of a word or token in a sequence. We evaluate four position embedding methods in this study: Absolute Position Embedding (APE)\cite{vaswani@attn}, Relative Position Embedding (RPE)\cite{shaw2018self}\cite{huang2020improve}, Rotary Position Embedding (RoPE)\cite{su2021roformer}, and Attention with Linear Biases (ALiBi)\cite{press2021train}.

Our analysis shows that when training on data with skewed position distribution, models are biased towards the first positions of a sequence, and the performance drops as the position of the word increases beyond what the model is trained on. In order to mitigate this problem, we proposed two batch-processing methods that can be implemented during the training process, \textbf{Random Position Perturbation} and \textbf{Context Perturbation} showed that we can mitigate those edge cases performance drops making the model more robust without any architecture changes\footnote{https://github.com/mehdibenamorr/Token-Positional-Bias}.
The main contributions of this study are to highlight an important issue in the application of LMs to NER and POS tagging and offer practical solutions to improve the performance of these models on these tasks. In light of this, we make the following contribution:
\begin{itemize}
    \item \textbf{Defining position bias} in the context of token classification and demonstrating its impact on Transformer LMs.
    \item Analyzing the impact of \textbf{class position distribution on the classification performance} on NER and POS tagging and four benchmarks. We show, that models trained on data with skewed position distribution are biased with performance drops for out-of-distribution tokens regardless of the position embedding method.
    \item Proposing two debiasing methods, \textbf{Random Position Perturbation} and \textbf{Context Perturbation}, to mitigate position bias and improve the model performance.
\end{itemize}

The rest of this paper is organized as follows: related work is presented in Section \ref{sec:related_work}. We define position bias and highlight its impact on NER and POS tagging in Section \ref{sec:analysis}. The two methods for overcoming position bias and corresponding experimental results are presented in Section \ref{sec:overcoming}. We present the conclusions in Section \ref{sec:conclusion}.

\section{Related Work}
\label{sec:related_work}
In recent years, there has been a growing interest in the use of Language Models (LMs) for various natural language processing (NLP) tasks. These LMs have achieved state-of-the-art performance on various benchmarks for NER and POS tagging, but their success is often dependent on the quality and quantity of the training data. 

\subsection{Data Imbalance and biases}

Various kinds of \textit{biases of datasets} have been observed in several research studies on NLP tasks. Li et al. \cite{li2019dice} addressed the issue of data imbalance in tasks such as tagging and machine reading comprehension (MRC), where the ratio of positive to negative examples is significantly low. They proposed a combination of Trversky index-based (dice) loss and focal loss to mitigate the effect of what they called \textbf{easy} negative examples. Zhou et al.\cite{zhou-chen-2021-learning} discussed the problem of overfitting on noisy labels (due to annotation errors) in models for Information Extraction tasks like NER and Relation Extraction (RE). 

\subsection{Positional Embeddings in Transformers}
\label{sec:pos_embeddings}
"Attention is all you need" \cite{vaswani@attn} introduced the concept of the self-attention mechanism in Transformer networks. Those networks have an encoder, decoder, and encoder-decoder stacks that mainly consist of Multi-Head attention layers. The attention is defined as follows:

\begin{displaymath}
    Attention(Q,K,V) = softmax(\frac{QK^T}{\sqrt{d_k}})V
\end{displaymath}

Where the input consists of matrices of queries $Q$, keys $K$ of dimension $d_k$, and values $V$. According to Vaswani et al. \cite{vaswani@attn}, they add “positional embeddings” to the input embeddings in order to inject positional information into the model. The positional encoding can be learned or fixed \cite{gehring2017}.

In BERT \cite{devlin2018bert}, Devlin et al. used absolute position embedding (APE) for the input to the self-attention head. 

\begin{equation}
    x_i = t_i + s_i + p_i 
\end{equation}

where $x_i$, $i$ $\in$ $\{0,\cdot,n-1\}$ is the input to the first transformer layer, $t_i$, $s_i$, and $p_i$ are the token, segments, and position embeddings respectively.

Shaw et al. \cite{shaw2018self} proposed a new approach to add position information in Transformers by extending the self-attention mechanism to take consideration of relative positions (RPE-key) between elements in a sequence. They introduce an edge representation, $a_{ij}$, into the attention computation to represent how a token $t_i$ attends to a token $t_j$. 
Dai et al. \cite{dai-etal-2019-transformer} proposed the Transformer-XL that enables learning context dependency beyond a fixed length without disrupting "temporal coherence". Transformer-XL is based on a segment-level recurrence mechanism and a new positional encoding scheme that captures longer-term dependency.
Huang et al. \cite{huang2020improve} argued that existing work on position encodings in Transformer architectures does not fully utilize position information and proposed four new methods that incorporate relative position in the self-attention query and key (RPE key-query) in different ways.

Su et al. \cite{su2021roformer} introduced a new method that encodes absolute position with rotation operations (RoPE), which allows the model to attend to relative positions better and enables the flexibility of sequence length and so on.

Wang et al. \cite{wang-chen-2020-position} conducted an empirical study on position embeddings of pre-trained Transformers to highlight whether position embeddings truly capture the meaning of positions and how they affect the performance of Transformers on NLP tasks. They provided an analysis of pre-trained position embeddings for some NLP tasks. They also introduce guiding tips on choosing suitable positional encoding depending on the task. 
\subsection{Position Bias}
\textit{Position Bias} has also been investigated in various problem definitions. In particular, Ko et al.\cite{ko2020look} analyzed position bias in the QA benchmark SQUAD\cite{rajpurkar2016squad}. They show that models predicting the position of answers can be biased in case correct answers primarily appear in a specific sentence position. To mitigate this bias, they proposed the use of word-level and sentence-level answer positions prior as a debiasing method. Hofstätter et al. \cite{hofstatter2021mitigating} further explored this phenomenon in two popular Question Answering datasets used for passage re-ranking and proposed a debiasing method for retrieval datasets to overcome favoring earlier positions inside passages. On the other hand, Ma et al. \cite{ma2021exploiting} highlighted the issue of robustness in state-of-the-art aspect sentiment classification (ASC) models and proposed that position bias, the importance of words closer to a concerning aspect, can be exploited for building more robust ASC models.

\section{Analysis}
\label{sec:analysis}
In this section, we conduct an analysis of position bias in benchmarks for NER and POS tagging. We begin by providing statistics that highlight the presence of position bias in those benchmarks. We then evaluate the impact of this bias on the performance of several baselines.

\subsection{Position Bias in Datasets}
\label{sec:data_analysis}
We analyze the following benchmarks that we will be using for all evaluations for both NER and POS tagging tasks.

\begin{itemize}
    \item \textbf{NER}: CoNLL03 \cite{sang2003introduction} is a widely used benchmark taken from the Reuters new corpus. Despite its small size, we chose it given the quality of its fine-grained annotations. On the other hand, OntoNotes5.0 \cite{weischedel2012ontonotes} is a large-scale dataset that has a wider range of named entities. 
\end{itemize}

\begin{itemize}
    \item \textbf{POS tagging}: The Universal Dependencies English Web Tree (UD\_en) \cite{silveira14gold} is a dependency treebank built over the source material of the English Web Treebank using the standardized Universal Dependencies' (UD) annotation scheme. The TweeBank \cite{liu-etal-2018-parsing} is a set of tweets parsed and manually annotated with the standard UD scheme. It is larger than UD\_en and has unique characteristics due to tweets having their own grammar and vocabulary.
\end{itemize}

\begin{figure}[t!]
    \centering
    \includegraphics[width=\linewidth]{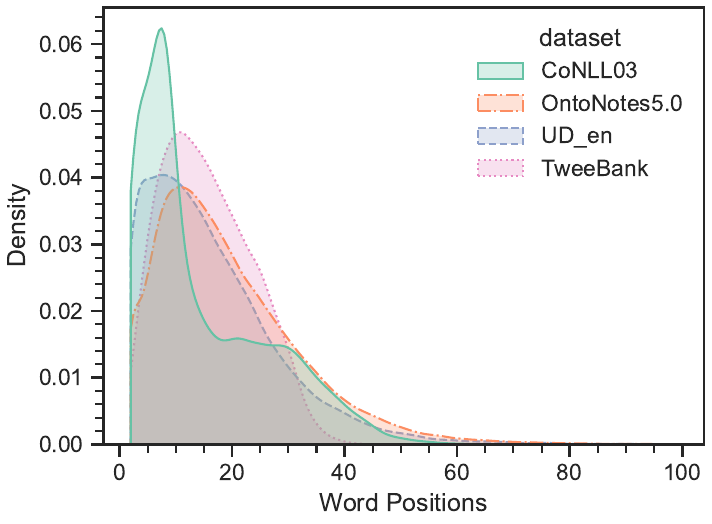}
    \caption{Density estimate of the distribution of sequence lengths for all benchmarks.}
    \label{fig:data_seq_length}
\end{figure}
In Figure \ref{fig:data_seq_length}, we look more into details of the distribution of sequence lengths across the four benchmarks. The probability density estimation shows a long tail distribution of sequences lengths. A key observation is the fact that CoNLL03 \cite{sang2003introduction} and UD\_en \cite{silveira14gold} have most of their examples in the range of [1,10]. OntoNotes5.0 \cite{weischedel2012ontonotes} and TweeBank \cite{liu-etal-2018-parsing} exhibit more balanced distributions for sequences shorter than 40 words than the other two.

\begin{table}[t!]
  \begin{tabular}{cccc}
    \toprule
    Dataset & \textbf{$\leq$25} & \textbf{$\geq$50} & \textbf{\#Seq}\\
    \midrule
    \texttt{CoNLL03} \cite{sang2003introduction} & 80\% & 3.5\% & 20.7k \\
    \texttt{OntoNotes5.0 \cite{weischedel2012ontonotes}}& 74\% & 7\% & 143.7k\\
    \midrule
    \texttt{UD\_en \cite{silveira14gold}}& 82\% & 5\% & 16.6k\\
    \texttt{TweeBank \cite{liu-etal-2018-parsing}}& 86\% & 0.1\% & 3.5k\\
    \bottomrule
  \end{tabular}
  \vspace{0.5em}
  \caption{Percentages from Sequence Length distribution in NER and POS tagging Benchmarks.}
  \label{tab:ner_stats}   
\end{table}

Table \ref{tab:ner_stats} shows overall frequencies of sequence length in the NER and POS tagging benchmarks. Intuitively, the length of training text sequences directly correlates with the position bias of words. As shown in the table, a large proportion of the sequences fall within a specific range of short lengths. For instance, 80\% of sequences in CoNLL03, 74\% of sequences in OntoNotes5.0, 82\% of sequences in UD\_en, and 86\% of sequences in TweeBank are 25 words or shorter. This highlights the potential for position bias to favor words in the early positions in longer sequences, given that only a small proportion of examples in these datasets are longer than 50 words. Notably, the smallest being TweeBank and CoNLL03 with 0.1\% and 3.5\%, respectively. Our assumption here is that if the model is mainly trained on short sequences, it might lead to the model being biased to the starting part of a long sequence in a testing scenario.

 \begin{figure}[t!]
    \centering
    \includegraphics[width=0.8\linewidth]{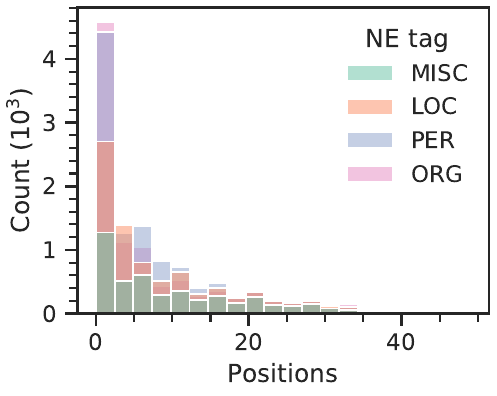}
    \caption{Class position distribution for Named Entity tags \textit{LOC}, \textit{ORG}, \textit{PER}, and \textit{MISC} in the training set for CoNLL03.}
    \label{fig:conll03_cls}
\end{figure}

\begin{figure}[t!]
    \centering
    \includegraphics[width=0.9\linewidth]{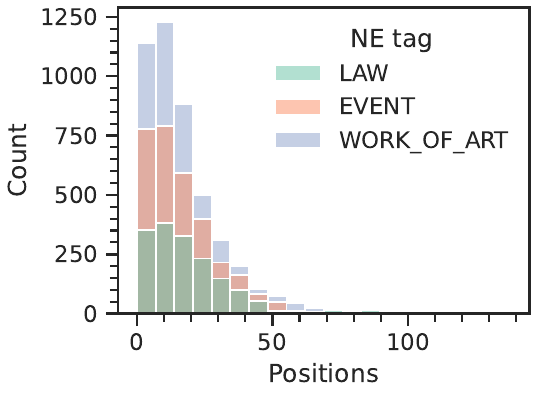}
    \caption{Class position distribution for NE tags \textit{EVENT}, \textit{WORK\_OF\_ART}, and \textit{LAW} for OntoNote5.0. }
    \label{fig:ontonote5.0_cls}
\end{figure}

Another aspect to analyze position bias in those benchmarks is the position distribution of a tag or a class appearance within a sequence. Figures \ref{fig:conll03_cls} and \ref{fig:ontonote5.0_cls} depict examples of the position distributions of named entities in the CoNLL03 and OntoNotes5.0 datasets, respectively. Both distributions of \textit{PER} and \textit{MISC} in Figure \ref{fig:conll03_cls} exhibit right-skewness, indicating that these named entities are more likely to appear at the start of the training examples, i.e, within the first 10 words. The skewness for the \textit{PER} class is more severe, with more than half of the instances appearing within the first 5 words. All three distributions in Figure \ref{fig:ontonote5.0_cls} have a wider spread within the first 40 words. The distribution for the \textit{LAW} class is the most evenly distributed. However, it is notable that it has the lowest frequency among the three classes. 

\begin{figure}[t!]
    \centering
    \includegraphics[width=0.8\linewidth]{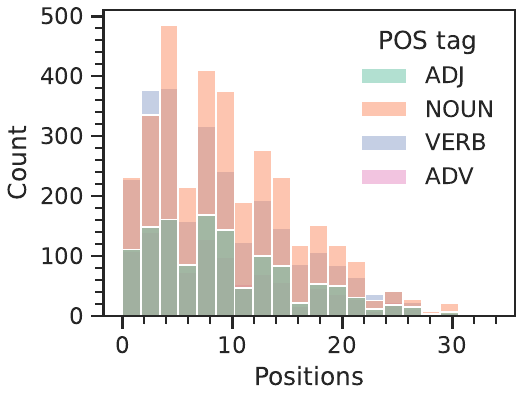}
    \caption{Class position distribution for POS tags \textit{ADJ}, \textit{NOUN}, \textit{VERB}, and \textit{ADV} in the training set for TweeBank. }
    \label{fig:tweebank_cls}
\end{figure}

For the POS tagging dataset TweeBank, we highlight the position distribution of the four most frequent POS tags in Figure \ref{fig:tweebank_cls}. All four tags have an even spread of positions, despite appearing within the first 25 words.

\textbf{In summary}, sequence length distribution in NER and POS benchmarks and class position within those sequences can be an indicator of a potential bias towards words in the early positions (e.g., from 1 to 10) of longer sequences when evaluating Transformer models.

\subsection{Impact of Position Bias}
\label{sec:position_bias}
To analyze the impact of the position bias on the performance of LMs in the downstream task of the NER and POS tagging, we first establish the experimental design of this evaluation. 

In our experiments, we choose BERT \cite{devlin2018bert} as the backbone model for evaluating the impact of position bias on the aforementioned tasks. BERT, as well as other non-recurrent self-attention based LMs, relies on  position embeddings to model dependencies between words in an input sequence. By learning those position embeddings that encode absolute positions in sequences, denoted as \textit{position IDs} that goes from 1 up to maximum length $M$ (i.e., 512 for BERT), token embeddings are then summed with position embedding to get the input sequence embeddings. As mentioned in Section \ref{sec:pos_embeddings}, Shaw et al. \cite{shaw2018self} and Huang et al. \cite{huang2020improve} introduced two improvements, noted as \textit{relative key} and \textit{relative key query} positional encoding respectively, that we will evaluate as one of the baselines.\\

\textbf{Baselines}: 
For all experiments, we modify the PyTorch implementation of the models\footnote{https://github.com/huggingface/transformers}. All models are trained with a training batch size of 16 and an evaluation batch size of 64 for 5 epochs. Every experiment is repeated with 5 random seeds [23456, 34567, 45678, 56789, 67890]. All model-related hyperparameters follow the default values used in the original work. The following models are used as baselines for comparison:
\begin{itemize}
    \item \textbf{BERT} \cite{devlin2018bert}: Base BERT encoder transformer model with \textbf{APE}.
    \item \textbf{ELECTRA} \cite{clark2020electra}: pre-trained as a discriminator instead of a generator, i.e., Masked Language Modeling (\textbf{APE}).
    \item \textbf{ERNIE} \cite{zhang-etal-2019-ernie}: pre-trained with additional knowledge information (\textbf{APE}).
    \item \textbf{BERT}\_\textit{key} \cite{shaw2018self}, \textbf{BERT}\_\textit{key-query}\cite{huang2020improve} with \textbf{RPE}.
    \item \textbf{RoFormer} \cite{su2021roformer}: BERT model trained with \textbf{RoPE}.
    \item \textbf{GPT2} (\textbf{APE}) \cite{radford2019language}, \textbf{BLOOM} (\textbf{ALiBi}) \cite{scao2022bloom}: decoder transformer models for text generation tasks.
\end{itemize}

 \begin{figure*}[h!]
    \centering
    \includegraphics[width=0.8\linewidth]{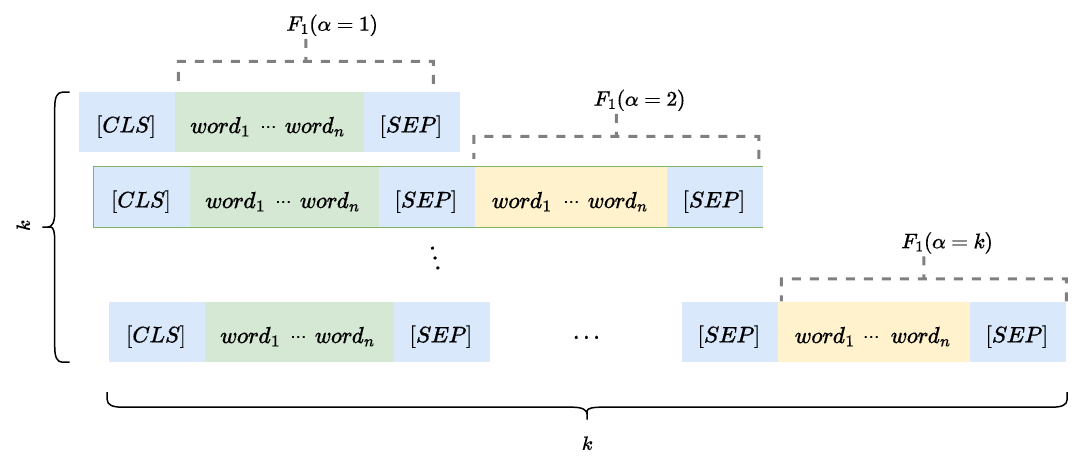}
    \caption{Visual of the proposed evaluation approach. Every example in the test set is duplicated $k$ \textbf{times}. The performance of the model is evaluated over a moving window $\alpha=\{1,\cdots,k\}$ for each test sample.}
    \label{fig:example_eval_approach}
\end{figure*}

\textbf{Approach}:
We propose a novel evaluation approach that involves \textbf{duplicating} sequences in the test set of the four benchmarks. 
The intuition behind this approach is that it allows us to study the effect of position bias on the model's performance without altering the context of the sentence. By repeating the same sentence $k$ times, we can observe how the model performs as the position of the named entities and POS tags within the new input sequence change. The assumption is that the local context of the original sentence is not substantially altered, which is crucial for evaluating the model's performance. If we were to concatenate different sentences, it would be difficult to disentangle the effect of position bias from the effect of noisy context on the performance.

For this experiment, we use a subset $\mathcal{D}^*$ of the datasets where we take all examples that fall between $1^{st}$ and $3^{rd}$ quantiles ($Q1$, $Q3$) of the sequence length distributions to exclude outliers of short or long examples and have more representative sets. We train on training sets $\mathcal{D}^*_{train}$ and evaluate the model on synthetic test data $\mathcal{D}^*_{test}(k)$, where $k$ is defined as the duplication coefficient. Please note that a separation token \textit{[SEP]} is added between repeated sentences since that was how sentences were pre-processed during the pre-training of BERT\cite{devlin2018bert}.
\begin{displaymath}
  \mathcal{D}^*_{test}(k) : \left\{ x^* \longleftarrow \underbrace{x, \dots , x}_{k\ \textrm{times}} \  \textrm{for}\   x \in \mathcal{D}^*_{test}\right\}
\end{displaymath}

Moreover, we vary the duplication factor $k$ for $\mathcal{D}^*_{test}(k)$ from 1 to 10. For $k=1$ it is simply the original test set, where the performance is reported in \textbf{F1} score. For $k \in \{2,\cdots,10\}$, we repeat the evaluation process and report the average and standard deviation for $F_1(\alpha)$ where $\alpha \in \{1,\cdots, k\}$ across all values of $k$ (see Figure. \ref{fig:example_eval_approach}).\\


\textbf{Metrics}:
In addition to the token classification evaluation metrics, we also introduce a position-based performance measure. We define the performance measure $F_1(\alpha)$ as the $F_1$ score based on a moving window of size $l_t$, where $\alpha$ is the position of the original subset within the duplicated sequence $x^*$ that ranges from 1 to $k$ and $l_x$ is the original length of the token sequence $x$. For each position of $\alpha$, the performance is evaluated on the subset of test set $\mathcal{D}^*_{test}(k,\alpha)$ using the following equation:

\begin{equation}
F_1(\alpha) = F_1(\mathcal{D}^*_{test}(k,\alpha)) \ ; \ \alpha \in \{1,\dots,k\}
\end{equation}

Where $\mathcal{D}^*_{test}(k,\alpha)$ is the subset of the duplicated test set $\mathcal{D}^*_{test}(k)$ where the window is moved to position $\alpha$. \\
This measure allows us to evaluate the performance of the model on different positions within the sentence, which provides more insights into the impact of position bias on the performance. Please note that following the same definition, we report $Prec(\alpha)$ and $Rec(\alpha)$ for the position-based precision and recall measures.

\begin{table*}[t!]
  \fontsize{6.5pt}{7.25pt}\selectfont
  \caption{Evaluation results for NER datasets, CoNLL03 and OntoNotes5.0. \textbf{F1}, F1(5), F1(10) are the f1 score on the original test set ($k=1$), and the position-based f1 score for $\alpha \in \{5,10\}$ respectively for $k=5,\cdots,10$. \underline{Bolded} values represent the lowest performance for each model.}
  \label{tab:ner_eval}
  \begin{tabular}{lllccc|ccc}
    \toprule
    \multicolumn{3}{c}{} & \multicolumn{3}{c}{\textbf{CoNLL03}} & \multicolumn{3}{c}{\textbf{OntoNotes5.0}} \\
    \midrule
    \textbf{APE} & & & \textbf{F1} & \textbf{F1(5)} & \textbf{F1(10)} & \textbf{F1} & \textbf{F1(5)} & \textbf{F1(10)}\\
    BERT \cite{devlin2018bert} & & & 90.66 ($\pm$.18) & 89.04 ($\pm$.31) & \textbf{88.35} ($\pm$.31) & 83.89 ($\pm$.39) & 82.06 ($\pm$.54) & \textbf{81.76} ($\pm$.45)\\
    ELECTRA \cite{clark2020electra} & & & 91 ($\pm$.02)& 84.9 ($\pm$.64) & \textbf{83.91} ($\pm$.37)& 84.49 ($\pm$.17)& 79.3 ($\pm$.58)& \textbf{78.6} ($\pm$.04)\\
    ERNIE \cite{zhang-etal-2019-ernie} & & & 90.1 ($\pm$.02)& 87.6 ($\pm$.4) & \textbf{87.1} ($\pm$.3)& 84.65 ($\pm$.1)& 81.9 ($\pm$\textbf{1.1})& \textbf{81.3} ($\pm$\textbf{1.2})\\
    GPT2 \cite{radford2019language} & & & 79.16 ($\pm$.1)& 57.3 ($\pm$\textbf{7}) & \textbf{43.3} ($\pm$\textbf{11})& 63.6 ($\pm$.08)& 60.46 ($\pm$\textbf{2})& \textbf{53.21} ($\pm$\textbf{4})\\
    \midrule
    \textbf{RPE} & & & & & & & & \\
    BERT(\textit{key}) \cite{shaw2018self} & & & 90.47 ($\pm$.34) & 85.77 ($\pm$.7)& \textbf{85.53} ($\pm$.71)& 84.53 ($\pm$.01)& 79.37 ($\pm$.62)& \textbf{79} ($\pm$.2)\\
    BERT(\textit{key-query}) \cite{huang2020improve} & & & 90.32 ($\pm$.24)& 85.25 ($\pm$.56) & \textbf{85.04} ($\pm$.3)& 84.27 ($\pm$.03)& 79.41 ($\pm$.46)& \textbf{79.3} ($\pm$.14)\\
    \midrule
    \textbf{RoPE} & & & & & & & & \\
    RoFormer \cite{su2021roformer} & & & 85.34 ($\pm$.28)& 82.7 ($\pm$0.9) & \textbf{81.38} ($\pm$1)& 79.81 ($\pm$.1)& 77.4 ($\pm$0.9)& \textbf{76.36} ($\pm$1)\\
    \midrule
    \textbf{ALiBi} & & & & & & & & \\
    BLOOM \cite{scao2022bloom} & & & 75.15 ($\pm$.12)& 70.87 ($\pm$1) & \textbf{68.8} ($\pm$.5)& 62.47 ($\pm$.1)& 61.44 ($\pm$\textbf{4.2})& \textbf{58.81} ($\pm$\textbf{5.6})\\
  \bottomrule
\end{tabular}
\end{table*}

\begin{table}[t!]
  \fontsize{6.5pt}{7.25pt}\selectfont
  \caption{Evaluation results for POS datasets, English UD and TweeBank. \textbf{F1}, F1(5), F1(10) are the f1 score on the original test set ($k=1$), and the position-based f1 score for $\alpha \in \{5,10\}$.}
  
  \label{tab:pos_eval}
  \begin{tabular}{lccc}
    \toprule
    \multicolumn{4}{c}{\textbf{UD\_en}}\\
    \midrule
    \textbf{APE}  & \textbf{F1} & \textbf{F1(5)} & \textbf{F1(10)} \\
    BERT & 95.73 ($\pm$.1) & 95.04 ($\pm$.03) & \textbf{95} ($\pm$.04)\\
    ELECTRA &  95.84 ($\pm$.01) & 94.32 ($\pm$.1) & \textbf{94.2} ($\pm$.08) \\
    ERNIE  &  96.29 ($\pm$.1) & 95.4 ($\pm$.12) & \textbf{95.4} ($\pm$.07) \\
    GPT2  &  90.78 ($\pm$.1) & 90 ($\pm$.2) & \textbf{88.2} ($\pm$.6) \\
    \midrule
    \textbf{RPE}  &  &  &  \\
    BERT(\textit{key})  & 95.96 ($\pm$.1) & 94.34 ($\pm$.19) & \textbf{94.3} ($\pm$.1)\\
    BERT(\textit{key query})  & 95.81 ($\pm$.05) & 94.6 ($\pm$.07) & \textbf{94.5} ($\pm$.01)\\
    \midrule
    \textbf{RoPE}  &  &  & \\
    RoFormer  & 91.87 ($\pm$.1) & 91.1 ($\pm$.1) & \textbf{90.86} ($\pm$.1) \\
    \midrule
    \textbf{ALiBi}  & & & \\
    BLOOM  & \textbf{89.03} ($\pm$.08) & 90.03 ($\pm$.66) & 89.36 ($\pm$.7) \\
    \midrule
    \multicolumn{4}{c}{\textbf{Tweebank}}\\
    \midrule
    \textbf{APE}  & \textbf{F1} & \textbf{F1(5)} & \textbf{F1(10)} \\
    BERT  & 78.63 ($\pm$.18) & 75.24 ($\pm$.79) & \textbf{74.4} ($\pm$.9) \\
    ELECTRA  & 68.97 ($\pm$.6) & 57.7 ($\pm$\textbf{3.8}) & \textbf{57.72} ($\pm$\textbf{4.})\\
    ERNIE  & 85.9 ($\pm$.03) & 81.7 ($\pm$\textbf{1}) & \textbf{81} ($\pm$\textbf{1.}) \\
    GPT2  & 79.28 ($\pm$.7) & 61.3 ($\pm$.5) & \textbf{48.7} ($\pm$.7) \\
    \midrule
    \textbf{RPE}  &  &  &  \\
    BERT(\textit{key}) & 81.8 ($\pm$.14) & 77.39 ($\pm$\textbf{1}) & \textbf{76.9} ($\pm$.8) \\
    BERT(\textit{key query})   & 80.45 ($\pm$.1) & 75.58 ($\pm$.5) & \textbf{75.32} ($\pm$.2)\\
    \midrule
    \textbf{RoPE}  &  &  & \\
    RoFormer  & 75.4 ($\pm$.2) & 72.48 ($\pm$.0.8) & \textbf{71.32} ($\pm$.0.8) \\
    \midrule
    \textbf{ALiBi}  & & & \\
    BLOOM & 84.9 ($\pm$.9) & 67.31 ($\pm$\textbf{2.7}) & \textbf{62.4} ($\pm$\textbf{2.}) \\
  \bottomrule
\end{tabular}
\end{table}

\subsection{Results}
 \begin{figure*}[t!]
    \centering
    \includegraphics[width=0.8\linewidth]{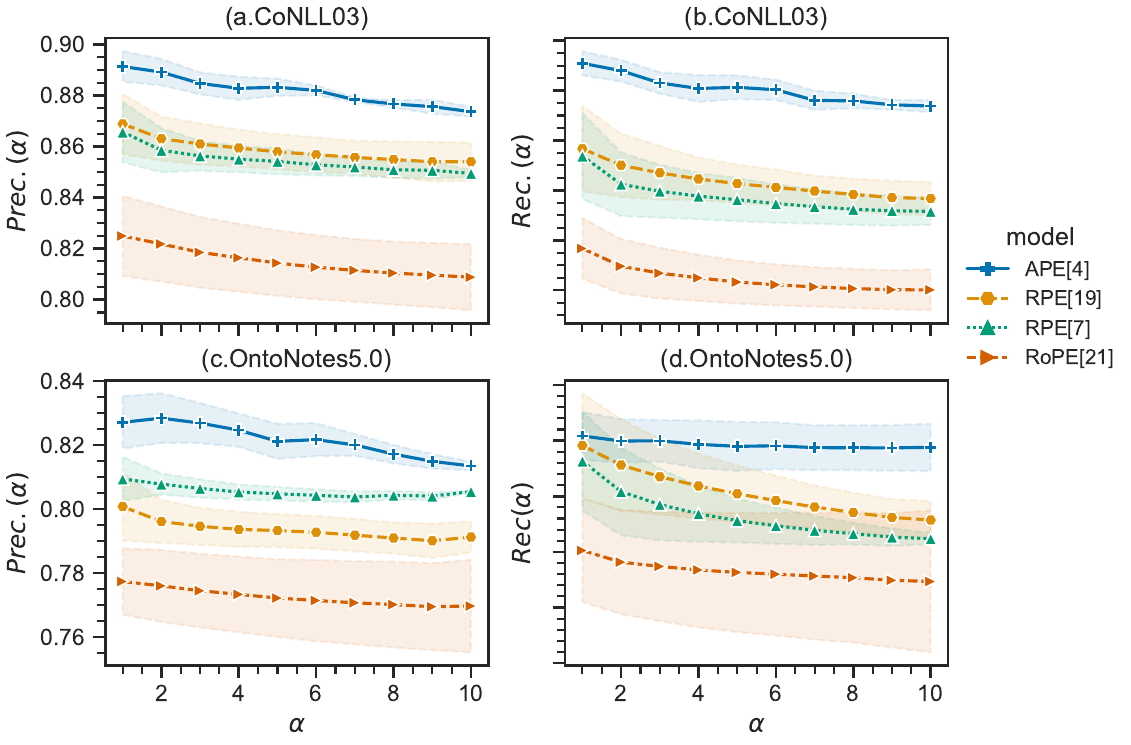}
    \caption{Evaluation results for $Prec(\alpha)$ and $Rec(\alpha)$ for the CoNLL03 ((a) and (b)) and OntoNotes5.0 ((c) and (d)) benchmarks.}
    \label{fig:overall_prec_recall}
\end{figure*}
\paragraph{\textbf{NER}}
Table \ref{tab:ner_eval} shows the position bias evaluation experimental results for CoNLL03 and OntoNotes5.0 datasets. We observe that all models have a drop in the F1 score when evaluated on the subset $\mathcal{D}^*_{test}(k,\alpha)$ with $\alpha$ being $\{5,10\}$ in this table. On CoNLL03, BERT has an F1 score of 90.66\% and F1(10) score of 88.35\%, which represents a drop of 3\% (at a standard deviation of .31), indicating a certain degree of position bias. Similarly, on the OntoNotes5.0 dataset, BERT has a drop of 2.13\% from an F1 score of 83.89\% when evaluated at position $\alpha=10$. BERT variants with \textit{relative key} and \textit{relative key query} positional encoding experienced a relatively higher drop of $\approx$ 5\% in F1 score for $\alpha=5$ and $10$ which might be partially correlated to the relative position embeddings being less robust to the \textbf{inductive property}, i.e., generalizing to longer sequences, on token classification tasks compared to the Question Answering (QA) \cite{huang2020improve}. RoFormer, BERT variant with RoPE,  demonstrated a drop of roughly 4\% in position-based F1 score from 85.3\% to 81.3\% at the 10-th position of the sequences. Figure \ref{fig:overall_prec_recall}.a and \ref{fig:overall_prec_recall}.b show a the same stable performance drop on CoNLL03 for BERT with different PE methods. In \ref{fig:overall_prec_recall}.c and \ref{fig:overall_prec_recall}.d, most PE models, except for APE, experienced a more severe degradation of the recall score $Rec(\alpha)$ while having a pseudo-stable precision performance. We interpret this as an of position bias on the ability of models to retrieve entities in later positions in longer sequences.

\paragraph{\textbf{POS tagging}}
Table \ref{tab:pos_eval} shows that for the UD\_en dataset, BERT and ERNIE performed the best with F1 scores of 95.73\% and 96.29\% respectively. However, when considering the other two measures, BERT dropped to 95.04\% and 95\% for F1(5) and F1(10) respectively. Similarly, ERNIE dropped to 95.4\% for F1(5) and F1(10). It is also worth noting that most models showed a low to negligible drop from $\alpha=5$ to $\alpha=10$. Overall, UD\_en shows a higher robustness in performance after the initial drop. 
For the TweeBank dataset, ERNIE performed the best with a score of 85.9\%. It is also the model with the highest robustness with the lowest decrease in F1(10) with only 4.9\% (at a standard deviation of 1) compared to the other models.

\paragraph{\textbf{Decoders}}
Furthermore, we evaluate decoder-only models (GPT2 and BLOOM in our case) to compare them to encoder models and show that they display similar behaviors when evaluated on $\mathcal{D}^*_{test}(k,\alpha)$. For CoNLL03, OntoNotes5.0, and Tweebank, we see a significant decline in the F1 score with the highest being on CoNLL03 for GPT2 with 36\%. Notably, the standard deviation values demonstrate higher values than the encoder-only models. This might suggest a high instability of decoder models in relation to position bias. We note that BLOOM was the only model to increase in performance on F1(5) and F1(10) during our evaluation. We report a slight increase of 1\% for F1(5) on UD\_en. However, this might be an outlier given the high standard deviation we saw from the decoder-only models.

\section{Overcoming Position Bias}
\label{sec:overcoming}
To address the issue of bias in language models, we introduce two methods that make adjustments to the input data during training. We will explain the underlying principles of each method, as well as evaluate their effectiveness. Overall, these techniques can be used to improve the robustness of Transformer LMs to token position bias.
\begin{table}[t!]
  \fontsize{6.5pt}{7.25pt}\selectfont
  \addtolength{\tabcolsep}{-4pt}
  \caption{Evaluation results of Random Position Perturbation and Context Perturbation on CoNLL03 and OntoNotes5.0. \underline{Underlined} values represent the best performance over each position windown $\alpha=1,5,10$.}
  \label{tab:ner_method}
  \begin{tabular}{lcccccc}
    \toprule
     & \multicolumn{3}{c}{\textbf{CoNLL03}} & \multicolumn{3}{c}{\textbf{OntoNotes5.0}} \\
    \midrule
    \textbf{Model}  & \textbf{F1} & \textbf{F1(5)} & \textbf{F1(10)} & \textbf{F1} & \textbf{F1(5)} & \textbf{F1(10)}\\
    BERT  & 90.66 ($\pm$.18) & 89.04 ($\pm$.31) & 88.35 ($\pm$.31) & 83.89 ($\pm$.39) & \underline{\textbf{82.06}} ($\pm$.54) & \underline{\textbf{81.76}} ($\pm$.45)\\
    BERT + RPP  & \underline{\textbf{90.91}} ($\pm$.05) & \underline{\textbf{90.23}} ($\pm$.22) & 89.64 ($\pm$.18) & \underline{\textbf{83.90}} ($\pm$.28) & 81.91 ($\pm$.45) & 81.42 ($\pm$.47)\\
    BERT + CP  & 90.88 ($\pm$.11) & 90.20 ($\pm$.1) & \underline{\textbf{90}} ($\pm$.18) & 83.82 ($\pm$.03) & 81.76 ($\pm$.32) & 81.46 ($\pm$.45)\\
    \toprule
    & \multicolumn{3}{c}{\textbf{UD\_en}} & \multicolumn{3}{c}{\textbf{Tweebank}} \\
    \midrule
    \textbf{Model}  & \textbf{F1} & \textbf{F1(5)} & \textbf{F1(10)} & \textbf{F1} & \textbf{F1(5)} & \textbf{F1(10)}\\
    BERT & 95.73 ($\pm$.1) & 95.04 ($\pm$.03) & 95 ($\pm$.04) & 78.63 ($\pm$.18) & 75.24 ($\pm$.79) & 74.4 ($\pm$.9) \\
    BERT + RPP & 95.88 ($\pm$.05) & 95.41 ($\pm$.07) & 95.30 ($\pm$.04) & 79.36 ($\pm$.12) & 77.25 ($\pm$.7) & 76.67 ($\pm$.6) \\
    BERT + CP & \underline{\textbf{95.96}} ($\pm$.03) & \underline{\textbf{95.61}} ($\pm$.06) & \underline{\textbf{95.7}} ($\pm$.14) & \underline{\textbf{79.69}} ($\pm$.3) & \underline{\textbf{78.28}} ($\pm$1.) & \underline{\textbf{78.12}} ($\pm$1.) \\
    \bottomrule
\end{tabular}
\end{table}

\subsection{Random Position Perturbation}

The attention mechanism in Transformers uses additional information, known as Positional Encoding, to give the model knowledge of the sequence order of the input tokens. This is done by adding positional embeddings to the word embeddings during the input process. Positional embeddings can be fixed or trained, with BERT using trained embeddings that learn an embedding for every position from 1 to the maximum length (512). The Random Position Perturbation (RPP) method aims to mitigate position bias by randomly shifting the position index $pos$ of each token in an input sequence by a factor $\tau$ during training. Given an input sequence $x:\{t,p\}$ in a training batch $\mathcal{B}_{train}$, the window $\tau$ is simply the distance between a newly selected position $p_r$ and the initial position of the first token $p_{0}=1$. It is worth noting that the \textit{[CLS]} token is at position 0. 

\begin{displaymath}
  \tau=p_r - 1 ; p_r \in I ;
\end{displaymath}

Where the new position $p_r$ is randomly selected from an interval $I=[l_t,M-l_t]$ of possible positions such that $l_t$ is the length of the token sequence $t$ and $M$ is the maximum length, i.e., $M=512$ for the BERT model, and $t_i$ is the i-th token in $t$.\\
We note the newly calculated position of token $t_i$ as $p_i^s$,
\begin{displaymath}
  p^s_{i} = p_{i} + \tau  \ \ \textrm{for}\ t_i \in t ;
\end{displaymath}

The resulting training batch is then expressed as follows,
\begin{displaymath}
  \mathcal{B}^*_{train} : \left\{ x:\{t, p^s\}  \ \textrm{for}\   x \in \mathcal{B}_{train}\right\}
\end{displaymath}

 \begin{figure}
    \centering
    \includegraphics[width=\linewidth]{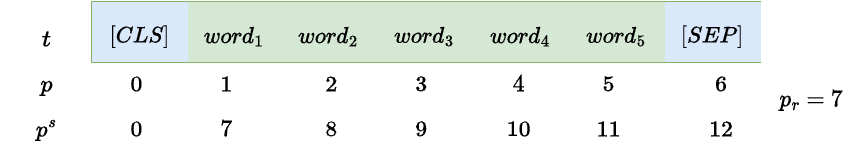}
    \caption{RPP with random position being $p_r=7$.}
    \label{fig:pos_shift}
\end{figure}

The intuition behind this method is that by randomly shifting the position of tokens, the model will learn to classify a token over an unbiased distribution of its positions within the input sequence. An example of the method output can be seen in Figure \ref{fig:pos_shift}.

\subsection{Context Perturbation}

The second proposed method is called Context Perturbation (CP). The method is based on selecting subsets of sequences $\mathcal{B}_{sub}$ in a training batch $\mathcal{B}_{train}$ such that the sum of their lengths is less or equal to the maximum length, $M=512$. For each subset of $n$ elements, the sequences are concatenated in $n$ random permutations of the order. This results in a batch of the same size $\mathcal{B}^*_{train}$, but with the sequences concatenated in different orders.

Formally, given a batch of input sequences $\mathcal{B}_{train}$, we split the batch into subsets of sequences such that $\sum len(x)<=M$ for $x \in \mathcal{B}_{sub}$ and $\cup_i \mathcal{B}^{(i)}_{sub} = \mathcal{B}_{train}$. Let the resulting concatenated sequences be denoted as $x_{concat}$. The context perturbation method is defined as:
\begin{displaymath}
   t_{concat} = t_{p_1} \oplus [SEP] \oplus t_{p_2} \oplus [SEP] ... \oplus t_{p_n} \oplus [SEP] ;
\end{displaymath}
With $(p_1, p_2, \cdots, p_n) \in \textrm{Set of} \ P_n^n=n! \ \textrm{arrangements}$ and $x_{p_i}$ being the i-th sequence in $\mathcal{B}_{sub}$ and $p_i$ being a random index from the set of $n$ sampled order permutations.

$\mathcal{B}^*_{train}$ is the resulting batch of same size as $\mathcal{B}_{train}$ with concatenated sequences in different random orderings.

By training on this perturbed batch, the model will learn to classify tokens within different positions and noisy contexts, thereby mitigating position bias and improving the performance.

\subsection{Results}

\paragraph{\textbf{NER}} Table \ref{tab:ner_method} presents the experimental results on NER benchmarks for the two methods. It can be observed that both the Random Position Perturbation (RPP) and Context Perturbation (CP) show improvement over the BERT (\textbf{APE}) baseline on CoNLL03. For RPP, we can see a significant increase of $\approx 1\%$ in F1($\alpha=5$) and F1($\alpha=10$) with a score of 90.23 and 89.64 compared to 89.04 and 88.35 for the base BERT model. For CP, F1($\alpha=10$) increased by almost 2\% keeping the model performance stable at around 90\% for all positions. Moreover, it is worth noting that the results on OntoNotes5.0 show no improvements for both methods. To investigate this, we looked into the precision and recall of some classes (see Appendix \ref{appendix}) that we highlighted their position distribution in Section \ref{sec:data_analysis}. 

\paragraph{\textbf{POS tagging}} The results in Table \ref{tab:pos_method} show that BERT with Random Position Perturbation (RPP) and Context Perturbation (CP) yield improved results on F1($\alpha=5$) and F1($\alpha=10$), respectively, compared to the baseline BERT model.
For the English UD dataset, the BERT + CP model achieved the best F1($\alpha=10$) score with 95.7 ($\pm$.14), while the BERT + RPP model achieved the best F1($\alpha=5$) score with 95.41 ($\pm$.07). Both models outperformed the baseline BERT model in terms of both metrics.
In the TweeBank dataset, the BERT + CP model showed the best performance with F1($\alpha=10$) of 78.12 ($\pm$1.), followed by the BERT + RPP model with F1($\alpha=5$) of 77.25 ($\pm$.7). Both models showed improved performance compared to the baseline BERT model.

Overall, both RPP and CP methods show the potential to reduce position bias for the NER task. While RPP shows a higher improvement in F1($\alpha=5$), CP shows a higher improvement in F1($\alpha=10$). Hence, the choice between the two methods can depend on the desired trade-off between improving position bias for the short-range and the long-range context.

\section{Conclusion}
\label{sec:conclusion}
Most of the position bias studies in the context of Transformer LMs have focused on downstream tasks such as QA and ASC and not token classification tasks, i.e., NER and POS tagging, to the best of our knowledge. Our position bias study shows that benchmarks of NER and POS tagging, like CoNLL03, OntoNotes5.0, UD\_en, and TweeBank, suffer from skewed position distribution of classes or tags. We show, with the proposed evaluation approach, the impact of this bias on Language models' performance when the token's position falls “out-of-distribution” for different Position Embedding methods. We introduce two computationally cheap and  easy-to-integrate methods to overcome position bias during training. 
Future work should include investigating attention scores and token representations to look into more tailored mitigation methods, and having a deeper analysis of the implicit relation between the position bias and sequence length.

\begin{acks}
This work is supported by
funds of the Federal Ministry of Food and Agriculture (BMEL) based on a decision of the Parliament of the Federal Republic of Germany. The Federal Office for Agriculture and Food (BLE) provides coordinating support for artificial intelligence (AI) in agriculture as funding organisation, grant number \textbf{2820KI012}. The work was also partly funded by BMBF under the funding code 01—S20049.
\end{acks}

\begin{figure*}
    \centering
    \includegraphics[width=.85\linewidth]{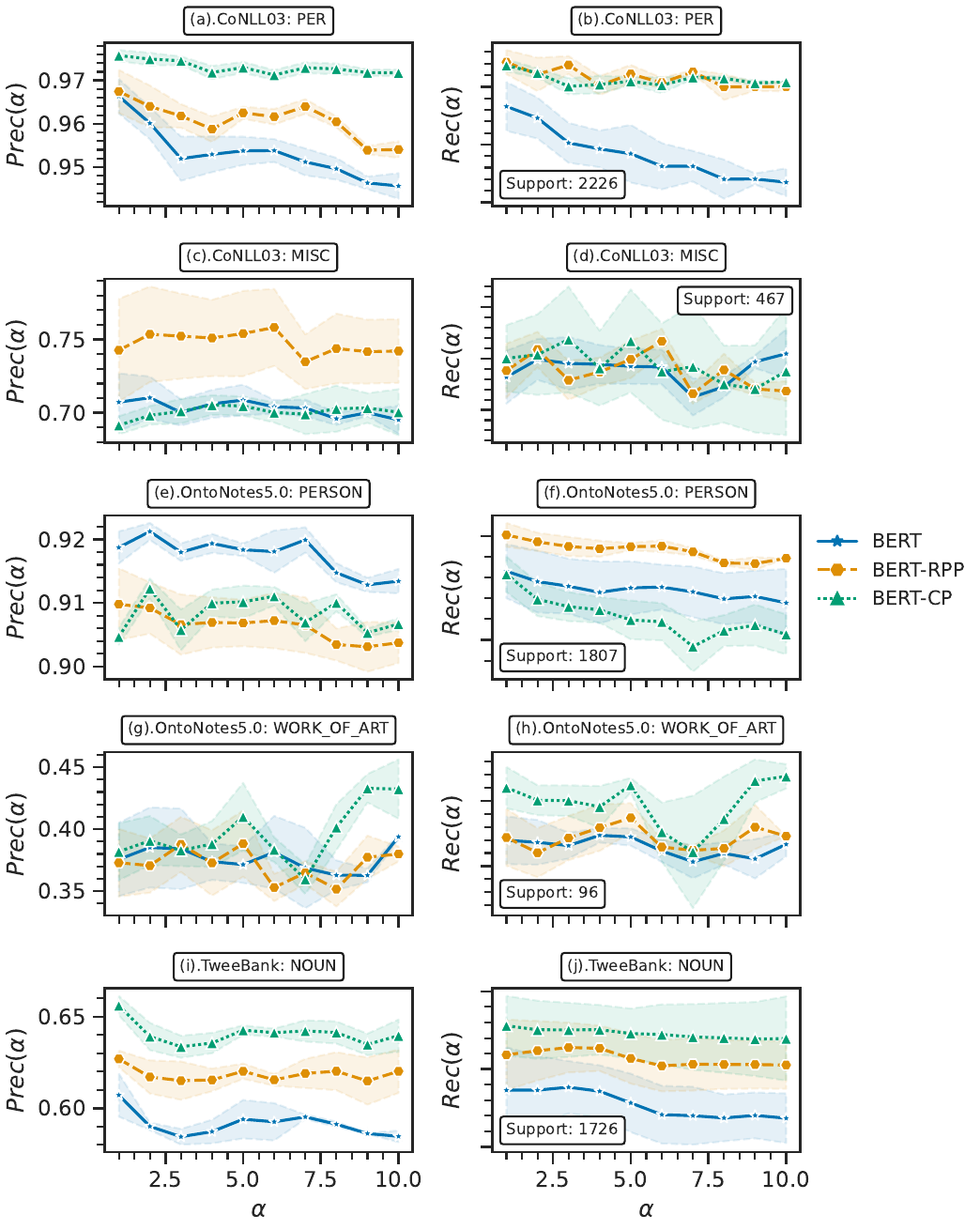}
    \caption{Class precision and recall scores ($Prec(\alpha)$, $Rec(\alpha)$) for $\alpha \in \{1,\cdots,10\}$. We compare the baseline BERT\cite{devlin2018bert} with the two methods RPP and CP. Only some classes with biased vs. balanced distribution as well as high vs. low presence in training set are highlighted.}
    \label{fig:indepth_perf}
\end{figure*}
\appendix
\section{Class Performances}
\label{appendix}
In Figures \ref{fig:indepth_perf} a, b, c, and d, we observe that both methods improved $Prec(\alpha)$ and $Rec(\alpha)$ for the biased class \textit{PER} more so than the better distributed \textit{MISC}. Nevertheless, we see for OntoNotes5.0 in Figures \ref{fig:indepth_perf}.e and f that for class \textit{PERSON}, which has a skewed distribution and well represented in the training data, the precision is worse than the baseline and the recall was only improved by the RPP method. This would explain the fact we saw similar performance after applying the two methods. This could be due to other factors that caused our nondeterministic approach for debiasing the training batches to influence those classes negatively.

\bibliographystyle{ACM-Reference-Format}
\bibliography{references}

\end{document}